\documentclass[11pt]{article}
\usepackage{acl2016}
\usepackage{pslatex}
\usepackage{amssymb}
\usepackage{latexsym}
\usepackage{amsmath}
\usepackage{graphicx}
\usepackage{tikz}
\usepackage{array,multirow}
\usepackage{url}
\usepackage{ulem}
\usepackage{rotating}
\usepackage{appendix}
\usepackage{verbatim}
\usepackage{color}
\usepackage{booktabs}
\usetikzlibrary{fadings}
\aclfinalcopy

\title{Neural Summarization by Extracting Sentences and Words}
\author{Jianpeng Cheng \qquad Mirella Lapata\\ILCC, School of Informatics, University of Edinburgh\\10 Crichton Street, Edinburgh EH8 9AB\\{\normalsize \tt jianpeng.cheng@ed.ac.uk \, mlap@inf.ed.ac.uk}}

\begin{document}
	\maketitle
	\begin{abstract}
          Traditional approaches to extractive summarization rely
          heavily on human-engineered features. In this work we
          propose a data-driven approach based on neural networks and
          continuous sentence features. We develop a general framework
          for single-document summarization composed of a hierarchical
          document encoder and an attention-based extractor. This
          architecture allows us to develop different classes of
          summarization models which can extract sentences or
          words. We train our models on large scale corpora containing
          hundreds of thousands of document-summary
          pairs\footnote{Resources are available for download at
          	\url{http://homepages.inf.ed.ac.uk/s1537177/resources.html}}. Experimental results on two summarization datasets
          demonstrate that our models obtain results comparable to the
          state of the art without any access to linguistic
          annotation.

	\end{abstract}
	
	%%%%%%%%%%%%%%%%%%%%%%%%%%%%%%%%%%%%%%%%%%%%%%%%%%%%%%%%%%%%%%%%%%%%%%%
	\section{Introduction}
	\label{sec:introduction}
	
	The need to access and digest large amounts of textual data has
	provided strong impetus to develop automatic summarization systems
	aiming to create shorter versions of one or more documents, whilst
	preserving their information content.  Much effort in automatic
	summarization has been devoted to sentence extraction, where a summary
	is created by identifying and subsequently concatenating the most
	salient text units in a document.

	Most extractive methods to date identify sentences based on
        human-engineered features. These include surface features such
        as sentence position and length \cite{radev2004mead}, the
        words in the title, the presence of proper nouns, content
        features such as word frequency
        \cite{nenkova2006compositional}, and event features such as
        action nouns \cite{filatova2004event}.  Sentences are
        typically assigned a score indicating the strength of presence
        of these features. Several methods have been used in order
        to select the summary sentences ranging from binary
        classifiers \cite{kupiec:ea:95}, to hidden Markov models
        \cite{conroy:oleary:2011}, graph-based algorithms
        \cite{erkan-radev:2004:EMNLP,mihalcea:2005:PosterDemo}, and
        integer linear programming \cite{woodsend2010automatic}.

	In this work we propose a data-driven approach to
        summarization based on neural networks and continuous sentence
        features.  There has been a surge of interest recently in
        repurposing sequence transduction neural network architectures
        for NLP tasks such as machine translation
        \cite{sutskever2014sequence}, question answering
        \cite{hermann2015teaching}, and sentence compression
        \cite{rush2015neural}. Central to these approaches is an
        encoder-decoder architecture modeled by recurrent neural
        networks. The encoder reads the source sequence into a list of
        continuous-space representations from which the decoder
        generates the target sequence. An attention mechanism
        \cite{bahdanau2014neural} is often used to locate the region
        of focus during decoding.
	
	We develop a general framework for single-document
        summarization which can be used to extract sentences or words.
        Our model includes a neural network-based hierarchical
        document reader or encoder and an attention-based content
        extractor. The role of the reader is to derive the meaning
        representation of a document based on its sentences and their
        constituent words. Our models adopt a variant of neural
        attention to extract sentences or words. Contrary to previous
        work where attention is an \emph{intermediate} step used to
        blend hidden units of an encoder to a vector propagating
        additional information to the decoder, our model applies
        attention \emph{directly} to select sentences or words of the
        input document as the output summary. Similar neural attention
        architectures have been previously used for geometry reasoning
        \cite{vinyals2015pointer}, under the name \textit{Pointer
          Networks}.
	
	One stumbling block to applying neural network models to
        extractive summarization is the lack of training data,
        i.e.,~documents with sentences (and words) labeled as
        summary-worthy. Inspired by previous work on summarization
        \cite{woodsend2010automatic,svore-vanderwende-burges:2007:EMNLP-CoNLL2007}
        and reading comprehension \cite{hermann2015teaching} we
        retrieve hundreds of thousands of news articles and
        corresponding highlights from the DailyMail
        website. Highlights usually appear as bullet points giving a
        brief overview of the information contained in the article
        (see Figure~\ref{example} for an example). Using a number of
        transformation and scoring algorithms, we are able to match
        highlights to document content and construct two large scale
        training datasets, one for sentence extraction and the other
        for word extraction. Previous approaches have used small scale
        training data in the range of a few hundred examples.
	
	Our work touches on several strands of research within summarization
	and neural sequence modeling. The idea of creating a summary by
	extracting words from the source document was pioneered in
	\newcite{bankoetal00} who view summarization as a problem analogous to
	statistical machine translation and generate headlines using
	statistical models for selecting and ordering the summary words. Our
	word-based model is similar in spirit, however, it operates over
	continuous representations, produces multi-sentence output, and
	jointly selects summary words and organizes them into sentences.  A
	few recent studies \cite{kobayashisummarization,yogatamaextractive}
	perform sentence extraction based on pre-trained sentence embeddings
	%. Different from
	%our approach, these works seek for extractive solutions 
	following an unsupervised optimization paradigm. Our work also
        uses continuous representations to express the meaning of
        sentences and documents, but importantly employs neural
        networks more directly to perform the actual summarization
        task.  

        \newcite{rush2015neural} propose a neural attention model for
        abstractive sentence compression which is trained on pairs of
        headlines and first sentences in an article. In contrast, our
        model summarizes documents rather than individual sentences,
        producing multi-sentential discourse. A major architectural
        difference is that our decoder selects output symbols from the
        document of interest rather than the entire vocabulary. This
        effectively helps us sidestep the difficulty of searching for
        the next output symbol under \textit{a large vocabulary}, with
        \textit{low-frequency words} and \textit{named entities} whose representations
        can be challenging to learn. \newcite{Gu:ea:16} and \newcite{gulcehre2016pointing} propose a similar ``copy'' mechanism in
        sentence compression and other tasks; their model can accommodate both
        generation and extraction by selecting which sub-sequences in
        the input sequence to copy in the output.

%During
%        training, our approach can be viewed as a variant of the
%        contrastive method for estimating the normalization constant
%        over the entire vocabulary. At test time, it renders the
%        search process for finding the next output symbol easier (and
%        faster) by focusing on the most probable document region.
	
	We evaluate our models both automatically (in terms of
        \textsc{Rouge}) and by humans on two datasets: the benchmark
        DUC~2002 document summarization corpus and our own DailyMail
        news highlights corpus.  Experimental results show that our
        summarizers achieve performance comparable to state-of-the-art
        systems employing hand-engineered features and sophisticated
        linguistic constraints.
	
	%%%%%%%%%%%%%%%%%%%%%%%%%%%%%%%%%%%%%%%%%%%%%%%%%%%%%%%%%%%%%%%%%%%%%%%
\begin{figure*}[t]
%	\begin{scriptsize}
		\begin{minipage}[t]{0.5\linewidth}
\begin{center}
			\begin{tabular}[t]{|@{~}p{15.5cm}@{~}|}
				\hline 
				%\vspace*{0.05ex}
				\textbf{AFL star blames vomiting cat for speeding} \\
				\uwave{Adelaide Crows defender Daniel Talia has kept his driving license,
					telling a court he was speeding 36km over the limit because he was
					distracted by his sick cat.}
				\\
				%\vspace*{-1.5ex}
				The 22-year-old AFL star, who drove 96km/h in a 60km/h road works
				zone on the South Eastern expressway in February, said he didn't see
				the reduced speed sign because he was so distracted by his cat
				vomiting violently in the back seat of his car.  \\ 
				%\vspace*{0.05ex}
				%\vspace*{-2ex}
				\uwave{In the Adelaide magistrates court on Wednesday, Magistrate Bob Harrap fined Talia \$824 for exceeding the speed limit by more than 30km/h.} \\
				He lost four demerit points, instead of seven, because
				of his significant training commitments. \\  \hline \hline 
				\begin{list}{\labelitemi}{\leftmargin=1em
						\itemsep=-.8ex} 
					\vspace*{-.45cm}\item \emph{Adelaide Crows defender Daniel Talia
						admits to speeding but says he didn't see road signs because his
						cat was vomiting in his car.}
					\item \emph{22-year-old Talia was fined \$824 and four demerit
						points, instead of seven, because of his 'significant' training
						commitments.} \vspace{-2ex}
				\end{list}\\ \hline
			\end{tabular}
\end{center}
		\end{minipage}
%	\end{scriptsize}
	\caption{DailyMail news article with highlights. Underlined sentences
		bear label~1, and~0 otherwise.}
	\label{example}
	\vspace*{-2.5mm}
\end{figure*}
	
	\section{Problem Formulation}
	\label{sec:problem-formulation}
	
	In this section we formally define the summarization tasks considered
	in this paper. Given a document~$D$ consisting of a sequence of
	sentences $\{s_1, \cdots, s_m\}$ and a word set $\{w_1, \cdots, w_n\}$, 
	we are interested in obtaining summaries
	at two levels of granularity, namely sentences and words.
	
	\textbf{Sentence extraction} aims to create a summary from~$D$
        by selecting a subset of~$j$ sentences
        (where~\mbox{$j<m$}). We do this by scoring each sentence
        within~$D$ and predicting a label~\mbox{$y_L \in {\{0,1\}}$}
        indicating whether the sentence should be included in the
        summary. As we apply supervised training, the objective is to
        maximize the likelihood of all sentence labels
        \mbox{$\mathbf{y}_L=(y_L^1, \cdots, y_L^m)$} given the input
        document~$D$ and model parameters $\theta$:
	\begin{equation}
		\log p(\mathbf{y}_L |D; \theta) = \sum\limits_{i=1}^{m} \log p(y_L^i |D; \theta)
		\label{sentence objective}
	\end{equation}
	
	Although extractive methods yield naturally grammatical
        summaries and require relatively little linguistic analysis,
        the selected sentences make for long summaries containing much
        redundant information.  For this reason, we also develop a
        model based on \textbf{word extraction} which seeks to find a
        subset of words\footnote{The vocabulary can also be extended
          to include a small set of commonly-used (high-frequency)
          words.}  in~$D$ and their optimal ordering so as to form a
        summary \mbox{$\mathbf{y}_s = (w'_1, \cdots, w'_k), w'_i \in
          D$}.  Compared to sentence extraction which is a sequence
        labeling problem, this task occupies the middle ground between
        full abstractive summarization which can exhibit a wide range
        of rewrite operations and extractive summarization which
        exhibits none. We formulate word extraction as a language
        generation task with an output vocabulary restricted to the
        original document. In our supervised setting, the training
        goal is to maximize the likelihood of the generated sentences,
        which can be further decomposed by enforcing conditional
        dependencies among their constituent words:
	\begin{equation}
		\hspace*{-.2cm}\log p(\mathbf{y}_s |D;
		\theta)\hspace*{-.1cm}=\hspace*{-.1cm}\sum\limits_{i=1}^{k}\hspace*{-.1cm}\log
		p(w'_i | D, w'_1,\hspace*{-.1cm}\cdots\hspace*{-.1cm}, w'_{i-1}; \theta)
		\label{word objective}
	\end{equation}
	In the following section, we discuss the data elicitation methods
	which allow us to train neural networks based on the above defined
	objectives.
	
	%%%%%%%%%%%%%%%%%%%%%%%%%%%%%%%%%%%%%%%%%%%%%%%%%%%%%%%%%%%%%%%%%%%%%%%

\section{Training Data for Summarization}
\label{sec:train-data-summ}
	
Data-driven neural summarization models require a large training
corpus of documents with labels indicating which sentences (or words)
should be in the summary.  Until now such corpora have been limited to
hundreds of examples (e.g., the DUC~2002 single document summarization
corpus) and thus used mostly for testing
\cite{woodsend2010automatic}. To overcome the paucity of annotated
data for training, we adopt a methodology similar to
\newcite{hermann2015teaching} and create two large-scale datasets, one
for sentence extraction and another one for word extraction.
	
In a nutshell, we retrieved\footnote{The script for constructing our
  datasets is modified from the one released in
  \newcite{hermann2015teaching}.}  hundreds of thousands of news
articles and their corresponding highlights from DailyMail (see
Figure~\ref{example} for an example). The highlights (created by news
editors) are genuinely abstractive summaries and therefore not readily
suited to supervised training. To create the training data for
\textbf{sentence extraction}, we reverse approximated the gold
standard label of each document sentence given the summary based on
their semantic correspondence
\cite{woodsend2010automatic}. Specifically, we designed a rule-based
system that determines whether a document sentence matches a highlight
and should be labeled with~1 (must be in the summary), and
0~otherwise. The rules take into account the position of the sentence
in the document, the unigram and bigram overlap between document
sentences and highlights, the number of entities appearing in the
highlight and in the document sentence.  We adjusted the weights of
the rules on 9,000 documents with manual sentence labels created by
\newcite{woodsend2010automatic}. The method obtained an accuracy
of~85\% when evaluated on a held-out set of 216~documents coming from
the same dataset and was subsequently used to label~200K documents.
Approximately 30\%~of the sentences in each document were deemed
summary-worthy.
	
For the creation of the \textbf{word extraction} dataset, we examine
the lexical overlap between the highlights and the news article. In
cases where all highlight words (after stemming) come from the
original document, the document-highlight pair constitutes a valid
training example and is added to the word extraction dataset. For
out-of-vocabulary (OOV) words, we try to find a semantically
equivalent replacement present in the news article. Specifically, we
check if a neighbor, represented by pre-trained\footnote{We used the
  Python Gensim library and the \mbox{300-dimensional}
  \emph{GoogleNews} vectors.} embeddings, is in the original document
and therefore constitutes a valid substitution. If we cannot find any
substitutes, we discard the document-highlight pair. Following this
procedure, we obtained a word extraction dataset containing~170K
articles, again from the DailyMail.

	%%%%%%%%%%%%%%%%%%%%%%%%%%%%%%%%%%%%%%%%%%%%%%%%%%%%%%%%%%%%%%%%%%%%%%%
\section{Neural Summarization Model}
\label{sec:model}
	
The key components of our summarization model include a neural
network-based hierarchical document reader and an attention-based
hierarchical content extractor. The hierarchical nature of our model
reflects the intuition that documents are generated compositionally
from words, sentences, paragraphs, or even larger units. We therefore
employ a representation framework which reflects the same
architecture, with global information being discovered and local
information being preserved.  Such a representation yields minimum
information loss and is flexible allowing us to apply neural attention
for selecting salient sentences and words within a larger context.  In
the following, we first describe the document reader, and then present
the details of our sentence and word extractors.%, which employs neural
%attention.

\subsection{Document Reader}
The role of the reader is to derive the meaning representation of the
document from its constituent sentences, each of which is treated as a
sequence of words. 
We first obtain representation vectors at the
sentence level using a single-layer convolutional neural network (CNN)
with a max-over-time pooling operation
\cite{blunsom2013recurrent,zhang2014chinese,kim2015character}. Next,
we build representations for documents using a standard recurrent
neural network (RNN) that recursively composes sentences. The CNN operates at the word level,
leading to the acquisition of sentence-level representations that are
then used as inputs to the RNN that acquires document-level
representations, in a hierarchical fashion. We describe these two
sub-components of the text reader below.
	
\paragraph{Convolutional Sentence Encoder}
	
We opted for a convolutional neural network model for representing
sentences for two reasons. Firstly, single-layer CNNs can be trained
effectively (without any long-term dependencies in the model) and
secondly, they have been successfully used for sentence-level
classification tasks such as sentiment analysis
\cite{kim2014convolutional}.
	%The reason for using a convolutional neural network for sentence
	%encoding is both for speed concern and to alleviate the vanishing
	%gradient problem caused by word-level long-term dependency.
Let~$d$ denote the dimension of word embeddings, and $s$~a document
sentence consisting of a sequence of $n$~words $(w_1, \cdots, w_n)$
which can be represented by a dense column matrix $\mathbf{W} \in
  \mathbb{R}^{n \times d}$. We apply a temporal narrow convolution
between $\mathbf{W}$ and a kernel $\mathbf{K} \in \mathbb{R}^{c \times
    d}$ of width $c$ as follows:
\begin{equation}
  \mathbf{f}^{i}_{j} = \tanh (\mathbf{W}_{j : j+c-1} \otimes  \mathbf{K} + b)
\end{equation}
where $\otimes$ equates to the Hadamard Product followed by a sum over
all elements. $\mathbf{f}^i_j $ denotes the $j$-th element of the
$i$-th feature map $\mathbf{f}^i$ and $b$ is the bias. We perform max pooling
over time to obtain a \emph{single} feature (the $i$th feature) representing the
sentence under the kernel $\mathbf{K}$ with width $c$:
\begin{equation}
	\mathbf{s}_{i, \mathbf{K}}= \max_j \mathbf{f}_j^i
\end{equation}
	
\begin{figure}[t!]
		\centering
		\includegraphics[width=0.48\textwidth]{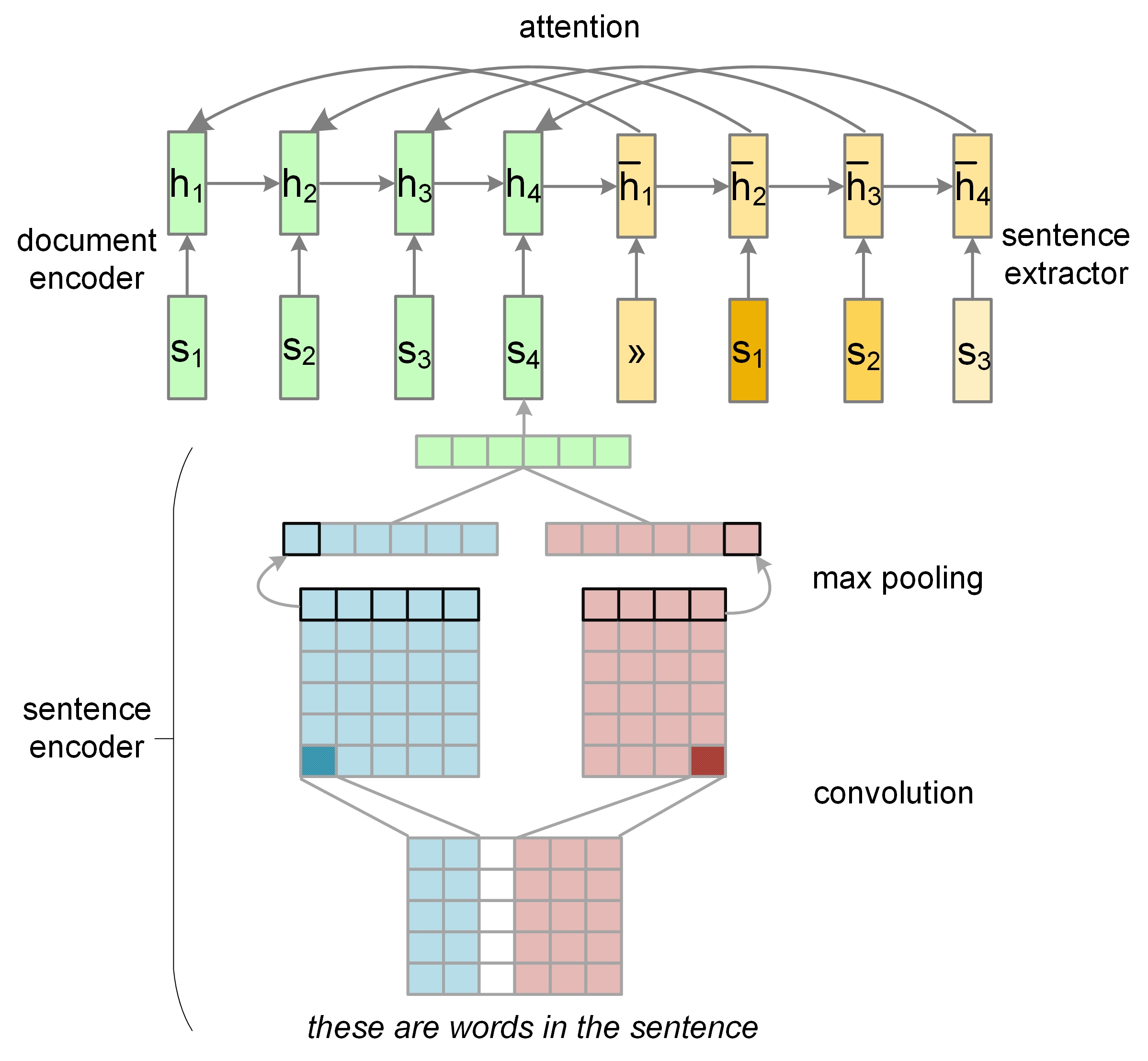}
                \caption{A recurrent convolutional document reader
                  with a neural sentence extractor.  }
		\label{sl}
		\vspace*{-2.5mm}
	\end{figure}

%This corresponds to one convolutional sentence feature obtained from
%one feature map. 
In practice, we use multiple feature maps to compute a list of
features that match the dimensionality of a sentence under each kernel width. In addition, we
apply multiple kernels with different widths to obtain a set of different sentence
vectors. Finally, we sum these sentence vectors to
obtain the final sentence representation. The CNN model is
schematically illustrated in Figure~\ref{sl} (bottom). In the example,
the sentence embeddings have six dimensions, so six feature maps are
used under each kernel width. The blue feature maps have width two and
the red feature maps have width three. The sentence embeddings
obtained under each kernel width are summed to get the final sentence
representation (denoted by green).
	
\paragraph{Recurrent Document Encoder}
At the document level, a recurrent neural network composes a sequence
of sentence vectors into a document vector. Note that this is a
somewhat simplistic attempt at capturing document organization at the
level of sentence to sentence transitions. One might view the hidden
states of the recurrent neural network as a list of partial
representations with each focusing mostly on the corresponding input
sentence given the previous context. These representations altogether
constitute the document representation, which captures local and
global sentential information with minimum compression.

The RNN we used has a Long Short-Term Memory (LSTM) activation unit
for ameliorating the vanishing gradient problem when training long
sequences \cite{hochreiter1997long}. Given a document \mbox{$d=(s_1,
  \cdots, s_m)$}, the hidden state at time step $t$, denoted by
$\mathbf{h_t}$, is updated as:
	\begin{equation}
		\begin{bmatrix}
			\mathbf{i}_t\\ \mathbf{f}_t\\ \mathbf{o}_t\\ \mathbf{\hat{c}}_t
		\end{bmatrix} =
		\begin{bmatrix} \sigma\\ \sigma\\ \sigma\\ \tanh
		\end{bmatrix} \mathbf{W}\cdot
		\begin{bmatrix} \mathbf{h}_{t-1}\\ \mathbf{s}_t
		\end{bmatrix}
	\end{equation}
	\begin{equation} \mathbf{c}_t = \mathbf{f}_t \odot \mathbf{c}_{t-1} +
		\mathbf{i}_t \odot \mathbf{\hat{c}}_t
	\end{equation}
	\begin{equation} \mathbf{h}_t = \mathbf{o}_t \odot \tanh(\mathbf{c}_t)
	\end{equation} where $\mathbf{W}$ is a learnable weight matrix. 
	Next, we discuss a special attention mechanism for extracting
	sentences and words given the recurrent document encoder just described,
	starting from the sentence extractor.
	
\subsection{Sentence Extractor}
In the standard neural sequence-to-sequence modeling paradigm
\cite{bahdanau2014neural}, an attention mechanism is used as an
intermediate step to decide which input region to focus on in order to
generate the next output. In contrast, our sentence extractor applies
attention to directly extract salient sentences after reading them.
	
The extractor is another recurrent neural network that labels
sentences sequentially, taking into account not only whether they are
individually relevant but also mutually redundant.  The complete
architecture for the document encoder and the sentence extractor is
shown in Figure~\ref{sl}. As can be seen, the next labeling decision
is made with both the encoded document and the previously labeled
sentences in mind. Given encoder hidden states~$(h_1, \cdots, h_m)$
and extractor hidden states~$(\bar{h}_1, \cdots, \bar{h}_m)$ at time
step~$t$, the decoder attends the~$t$-th sentence by relating its
current decoding state to the corresponding encoding state:
\begin{equation}
	\bar{\mathbf{h}}_{t} = \text{LSTM} ( p_{t-1} \mathbf{s}_{t-1}, \mathbf{\bar{h}}_{t-1})
	\label{update}
\end{equation}
\begin{equation}
  p(y_L(t)=1 | D ) = \sigma(\text{MLP} (\mathbf{\bar{h}}_t : \mathbf{h}_t) )
\end{equation}
where MLP is a multi-layer neural network with as input the concatenation of $\mathbf{\bar{h}}_t$ and $\mathbf{h}_t$. 
$p_{t-1}$ represents the degree to which the extractor believes
the previous sentence should be extracted and memorized ($p_{t-1}$=1 if the system
is certain; 0 otherwise).

%; it also represents the proportion
%of the sentence content that should go into the memory of the
%extractor.

In practice, there is a discrepancy between training and testing such
a model. During training we know the true label~$p_{t-1}$ of the
previous sentence, whereas at test time $p_{t-1}$ is unknown and has
to be predicted by the model. The discrepancy can lead to quickly
accumulating prediction errors, especially when mistakes are made
early in the sequence labeling process. To mitigate this, we adopt a
curriculum learning strategy \cite{bengio2015scheduled}: at the
beginning of training when $p_{t-1}$ cannot be predicted accurately,
we set it to the true label of the previous sentence; as training goes
on, we gradually shift its value to the predicted label~$p(y_L(t-1)=1
| d )$.

\begin{figure}[t!]
\centering
\includegraphics[width=0.5\textwidth]{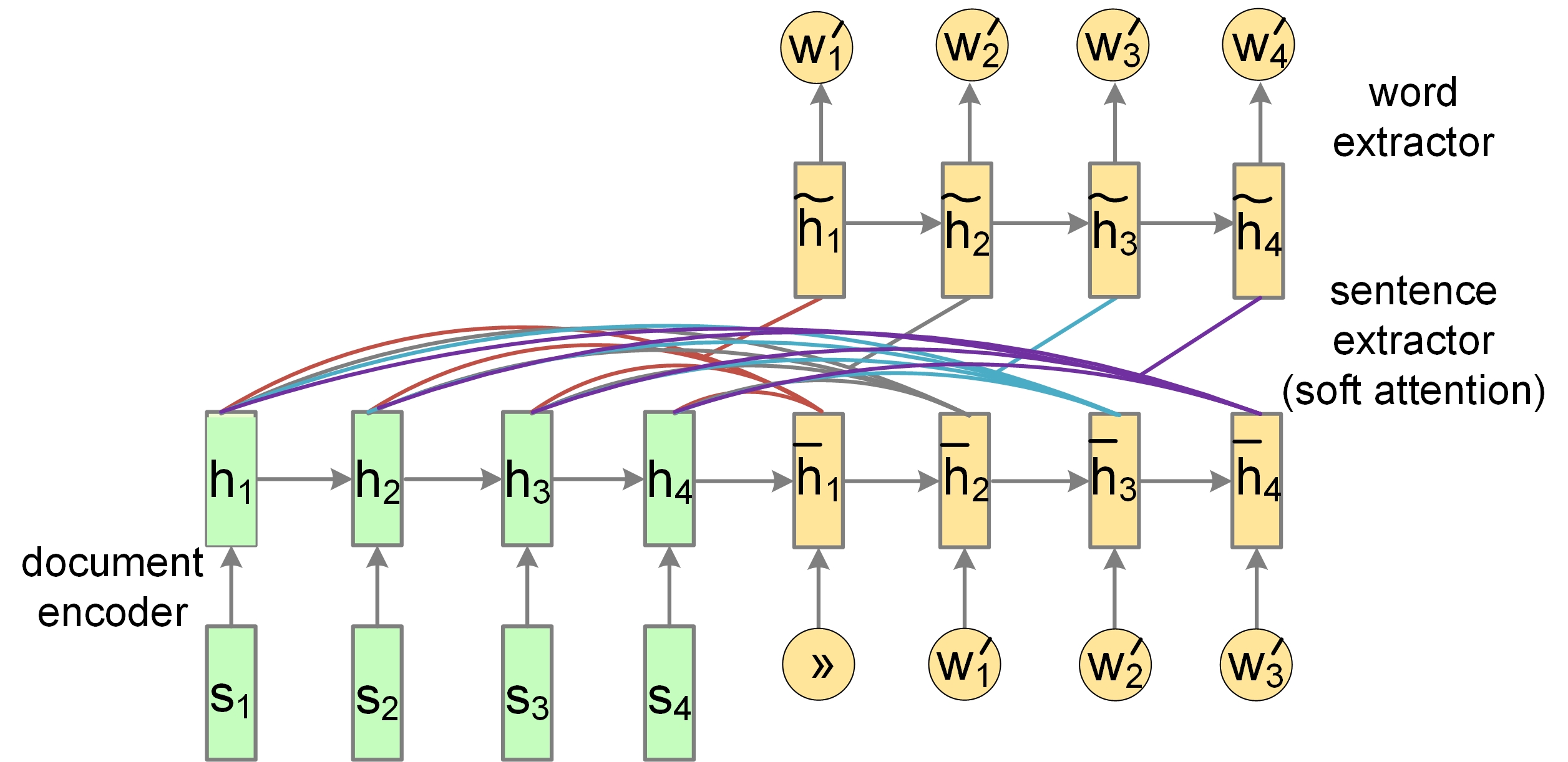}
\caption{Neural attention mechanism for word extraction.}
\label{we}
\end{figure}
	
\subsection{Word Extractor}
Compared to sentence extraction which is a purely sequence labeling
task, word extraction is closer to a generation task where relevant
content must be selected and then rendered fluently and
grammatically. A small extension to the structure of the sequential
labeling model makes it suitable for generation: instead of predicting
a label for the next sentence at each time step, the model directly
outputs the next word in the summary. The model uses a
\emph{hierarchical} attention architecture: at time step~$t$, the
decoder softly\footnote{A simpler model would use hard attention to
  select a sentence first and then a few words from it as a summary,
  but this would render the system non-differentiable for
  training. Although hard attention can be trained with the REINFORCE
  algorithm \cite{williams1992simple}, it requires sampling of
  discrete actions and could lead to high variance.}
%  during training.}
attends each document sentence and
subsequently attends each word in the document and computes the
probability of the next word to be included in the summary $p(w'_t = w_i|
d, w'_1, \cdots, w'_{t-1})$ with a softmax classifier: % as follows:

\begin{equation}
  \bar{\mathbf{h}}_{t} = \text{LSTM} ( \mathbf{w'}_{t-1},
  \mathbf{\bar{h}}_{t-1})\footnote{We empirically found that feeding
    the previous sentence-level attention vector as additional
    input to the LSTM would lead to  small performance improvements.
    This is not shown in the equation.} 
\end{equation}
\begin{equation}
	a_j^t = \mathbf{z}^\mathtt{T} \tanh(\mathbf{W}_e \mathbf{\bar{h}}_t + \mathbf{W}_r \mathbf{h}_j), h_j \in D
\end{equation}
\begin{equation}
	b_j^t = \text{softmax}(a_j^t)
\end{equation}
\begin{equation}
    \mathbf{\tilde{h}}_t = \sum\limits_{j=1}^{m} b_j^t \mathbf{h}_j
\end{equation}
\begin{equation}
u_i^t = \mathbf{v}^\mathtt{T} \tanh(\mathbf{W}_{e'} \mathbf{\tilde{h}}_t + \mathbf{W}_{r'} \mathbf{w}_i), w_i \in D
\end{equation}
\begin{equation}
p(w'_t  = w_i| D, w'_1, \cdots, w'_{t-1}) = \text{softmax}(u_i^t)
\end{equation}
In the above equations, $\mathbf{w}_i$ corresponds to the vector of the $i$-th
word in the input document, whereas $\mathbf{z}$, $\mathbf{W}_e$,
$\mathbf{W}_r$, $\mathbf{v}$, $\mathbf{W}_{e'}$, and $\mathbf{W}_{r'}$ are model
weights. The model architecture is shown in Figure~\ref{we}.

The word extractor can be viewed as a conditional language model with
a vocabulary constraint. In practice, it is not
powerful enough to enforce grammaticality due to the lexical diversity
and sparsity of the document highlights. A possible enhancement would
be to pair the extractor with a neural language model, which can be
pre-trained on a large amount of unlabeled documents and then
jointly tuned with the extractor during decoding
\cite{gulcehre2015using}.  A simpler alternative which we adopt is to
use $n$-gram features collected from the document to rerank candidate summaries
obtained via beam decoding. We incorporate the features in a
log-linear reranker whose feature weights are optimized with minimum
error rate training \cite{och2003minimum}.

\section{Experimental Setup}

In this section we present our experimental setup for assessing the
performance of our summarization models. We discuss the datasets used
for training and evaluation, give implementation details, briefly
introduce comparison models, and explain how system output was
evaluated.	
	
\paragraph{Datasets}
We trained our sentence- and word-based summarization models on the
two datasets created from DailyMail news. Each dataset was split into
approximately 90\%~for training, 5\%~for validation, and 5\%~for
testing.  We evaluated the models on the DUC-2002 single document
summarization task. In total, there are~567 documents belonging to~59
different clusters of various news topics. Each document is associated
with two versions of 100-word\footnote{According to the DUC2002 guidelines \url{http://www-nlpir.nist.gov/projects/duc/guidelines/2002.html}, the generated summary should be within 100 words.} manual summaries produced by human
annotators.  We also evaluated our models on~500 articles from the
DailyMail test set (with the human authored highlights as
goldstandard). We sampled article-highlight pairs so that the
highlights include a minimum of~3 sentences. The average byte count for each document is 278. As there is no established evaluation standard
for this task, we also report ROUGE evaluation on the entire DailyMail test set with varying limits. Please refer to the appendix for more information.
	
\paragraph{Implementation Details}
We trained our models with Adam \cite{kingma2014adam} with initial
learning rate 0.001. The two momentum parameters were set to 0.99 and
0.999 respectively. We performed mini-batch training with a batch size
of~20 documents. All input documents were padded to the same length
with an additional mask variable storing the real length for each
document. The size of word, sentence, and document embeddings were set
to 150, 300, and 750, respectively. For the convolutional sentence
model, we followed \newcite{kim2015character}\footnote{The CNN-LSTM
  architecture is publicly available at
  \url{https://github.com/yoonkim/lstm-char-cnn}.} and used a list of
kernel sizes \{1, 2, 3, 4, 5, 6, 7\}. For the recurrent document model
and the sentence extractor, we used as regularization dropout with
probability~0.5 on the LSTM input-to-hidden layers and the scoring
layer. The depth of each LSTM module was~1. All LSTM parameters were
randomly initialized over a uniform distribution within~\mbox{[-0.05,
  0.05]}. The word vectors were initialized with 150 dimensional
pre-trained embeddings.\footnote{We used the \textit{word2vec}
  \cite{mikolov2013distributed} skip-gram model with context window
  size~6, negative sampling size~10 and~hierarchical softmax~1. The
  model was trained on the \textit{Google 1-billion} word benchmark
  \cite{Chelba:ea:2014}.}
	
Proper nouns pose a problem for {embedding-based} approaches,
especially when these are rare or unknown (e.g.,~at test
time). \newcite{rush2015neural} address this issue by adding a new set
of features and a log-linear model component to their system.  As our
model enjoys the advantage of generation by extraction, we can force
the model to inspect the context surrounding an entity and its
relative position in the sentence in order to discover extractive
patterns, placing less emphasis on the meaning representation of the
entity itself. Specifically, we perform named entity recognition with
the package provided by \newcite{hermann2015teaching} and maintain a
set of randomly initialized entity embeddings.  During training, the
index of the entities is permuted to introduce some noise but also
robustness in the data. A similar data augmentation approach has been
used for reading comprehension \cite{hermann2015teaching}.

A common problem with extractive methods based on sentence labeling is
that there is no constraint on the number of sentences being selected
at test time. We address this by reranking the positively labeled
sentences with the probability scores obtained from the softmax layer
(rather than the label itself). In other words, we are more interested
in is the relative ranking of each sentence rather than their exact
scores. This suggests that an alternative to training the network
would be to employ a ranking-based objective or a \emph{learning to
  rank} algorithm. However, we leave this to future work. We use the
three sentences with the highest scores as the summary (also subject
to the word or byte limit of the evaluation protocol).

Another issue relates to the word extraction model which is
challenging to batch since each document possesses a distinct
vocabulary. We sidestep this during training by performing negative
sampling \cite{mikolov2013distributed} which trims the vocabulary of
different documents to the same length. At each decoding step the
model is trained to differentiate the true target word from 20~noise
samples. At test time we still loop through the words in the input
document (and a stop-word list) to decide which word to output next.

\paragraph{System Comparisons}
We compared the output of our models to various summarization methods.
These included the standard baseline of simply selecting the
``leading'' three sentences from each document as the summary.  We
also built a sentence extraction baseline classifier using logistic
regression and human engineered features. The classifier was trained
on the same datasets as our neural network models with the following
features: sentence length, sentence position, number of entities in
the sentence, sentence-to-sentence cohesion, and sentence-to-document
relevance. Sentence-to-sentence cohesion was computed by calculating
for every document sentence its embedding similarity with every other
sentence in the same document. The feature was the normalized sum of
these similarity scores.  Sentence embeddings were obtained by
averaging the constituent word embeddings.  Sentence-to-document
relevance was computed similarly. We calculated for each sentence its
embedding similarity with the document (represented as bag-of-words),
and normalized the score. The word embeddings used in this baseline
are the same as the pre-trained ones used for our neural models.

In addition, we included a neural abstractive summarization
baseline. This system has a similar architecture to our word
extraction model except that it uses an open vocabulary during
decoding. It can also be viewed as a hierarchical document-level
extension of the abstractive sentence summarizer proposed by
\newcite{rush2015neural}. We trained this model with
negative sampling to avoid the excessive computation of the
normalization constant.

Finally, we compared our models to three previously published systems
which have shown competitive performance on the DUC2002 single
document summarization task. The first approach is the phrase-based
extraction model of \newcite{woodsend2010automatic}. Their system
learns to produce highlights from parsed input (phrase structure trees
and dependency graphs); it selects salient phrases and recombines them
subject to length, coverage, and grammar constraints enforced via
integer linear programming (ILP). Like ours, this model is trained on
document-highlight pairs, and produces telegraphic-style bullet points
rather than full-blown summaries.  The other two systems,
\textsc{tgraph} \cite{parveen-ramsl-strube} and \textsc{urank}
\cite{wan2010towards}, produce more typical summaries and represent
the state of the art.  \textsc{tgraph} is a graph-based sentence
extraction model, where the graph is constructed from topic models and
the optimization is performed by constrained ILP.  \textsc{urank}
adopts a unified ranking system for both single- and multi-document
summarization.
	
\begin{table}[t]
%		\small
		\centering
		\begin{tabular}{|@{~}l@{\hspace{2ex}}|c@{\hspace{2ex}}c@{\hspace{2ex}}c@{~}|} \hline
			 \multicolumn{1}{|@{~}l|}{DUC 2002} &  \textsc{Rouge-1}  &  \textsc{Rouge-2}  & \textsc{Rouge-l} \\ \hline\hline
			 \textsc{lead} & 43.6& 21.0  & 40.2 \\
			 \textsc{lreg} & 43.8  & 20.7 & 40.3 \\
			 \textsc{ilp} & 45.4 & 21.3 & 42.8 \\
			 \textsc{nn-abs}  & 15.8 & \hspace*{.17cm}5.2  & 13.8\\ 
			 \textsc{tgraph} & 48.1 & \textbf{24.3} & ---\\
			 \textsc{urank} & \textbf{48.5} & 21.5 & ---\\
			 \textsc{nn-se} & 47.4 & 23.0   & \textbf{43.5}\\
			 %			\raisebox{1cm}[0pt]{\begin{sideways}DUC~2002\end{sideways}} 
			 \textsc{nn-we} & 27.0 & \hspace*{.17cm}7.9 & 22.8 \\
			\hline 
			\multicolumn{4}{c}{} \\\hline
			 \multicolumn{1}{|@{~}l|}{DailyMail} &
			 \textsc{Rouge-1}  & \textsc{Rouge-2}  & \textsc{Rouge-l} \\ \hline\hline
			 \textsc{Lead} & 20.4 &  7.7 & 11.4 \\
			 \textsc{lreg} & 18.5 & 6.9 & 10.2 \\ 
			 \textsc{nn-abs}  & \hspace*{.17cm}7.8 & 1.7 & \hspace*{.5em}7.1\\  
			 \textsc{nn-se} & \textbf{21.2} & \textbf{8.3} & \textbf{12.0}\\
			 %			\raisebox{.1cm}[0pt]{\begin{sideways}DailyMail\end{sideways}}
			 \textsc{nn-we} &  15.7 & 6.4& 9.8 \\\hline
		\end{tabular}
		\caption{\label{duc} \textsc{Rouge} evaluation (\%) on the DUC-2002 and 500 DailyMail samples.}
	\end{table}
	
\paragraph{Evaluation}
	
We evaluated the quality of the summaries automatically using
\textsc{Rouge} \cite{lin:03}.  We report unigram and bigram overlap
(\mbox{\textsc{Rouge-1,2}}) as a means of assessing informativeness
and the longest common subsequence (\mbox{\textsc{Rouge-L}}) as a
means of assessing fluency.
	
In addition, we evaluated the generated summaries by eliciting human
judgments for 20~randomly sampled DUC~2002 test documents.
Participants were presented with a news article and summaries
generated by a list of systems. These include two neural network
systems (sentence- and word-based extraction), the neural abstractive
system described earlier, the lead baseline, the phrase-based ILP
model\footnote{We are grateful to Kristian Woodsend for giving us
  access to the output of his system. Unfortunately, we do not have
  access to the output of \textsc{tgraph} or \textsc{urank} for
  inclusion in the human evaluation.} of
\newcite{woodsend2010automatic}, and the human authored
summary. Subjects were asked to rank the summaries from best to worst
(with ties allowed) in order of informativeness (does the summary
capture important information in the article?) and fluency (is the
summary written in well-formed
English?). %Each participant is required to rank
%\textit{all} the summaries. 
We elicited human judgments using Amazon's Mechanical Turk
crowdsourcing platform. Participants (self-reported native English
speakers) saw 2~random articles per session. We collected 5~responses
per document.

	\begin{figure*}[ht]
		\centering
		\scriptsize
		\begin{tabular}{|@{~}p{16cm}@{~}|}
                  \hline 
                  \textbf{sentence extraction}: \\
                  \tikz[baseline=(A.base)]{\node[opacity=0](A) {
                      a gang of at least three people poured gasoline on a car that stopped to fill up at \emph{entity5} gas station early on Saturday morning and set the vehicle on fire};
                    \shade[inner color=red!40] (A.south east) rectangle (A.north west);
                    \path (A.center) \pgfextra{\pgftext{a gang of at least three people poured gasoline on a car that stopped to fill up at \emph{entity5} gas station early on Saturday morning and set the vehicle on fire}};}
                  \\
                  \tikz[baseline=(A.base)]{\node[opacity=0](A) {the driver of the car, who has not been identified, said he got into an argument with the suspects while he was pumping gas at a \emph{entity13} in \emph{entity14}};
                    \shade[inner color=red!30] (A.south east) rectangle (A.north west);
                    \path (A.center) \pgfextra{\pgftext{the driver of the car, who has not been identified, said he got into an argument with the suspects while he was pumping gas at a \emph{entity13} in \emph{entity14}}};}
                  \\
                  \tikz[baseline=(A.base)]{\node[opacity=0](A) {the group covered his white \emph{entity16} in gasoline and lit it ablaze while there were two passengers inside};
                    \shade[inner color=red!10] (A.south east) rectangle (A.north west);
                    \path (A.center) \pgfextra{\pgftext{the group covered his white \emph{entity16} in gasoline and lit it ablaze while there were two passengers inside}};}
                  \\
                  at least three people poured gasoline on a car and lit it on fire at a \emph{entity14} gas station explosive situation\\ 
                  the passengers and the driver were not hurt during the incident but the car was completely ruined \\
                  \tikz[baseline=(A.base)]{\node[opacity=0](A) {the man's grandmother said the fire was lit after the suspects attempted to carjack her grandson, \emph{entity33} reported};
                    \shade[inner color=red!40] (A.south east) rectangle (A.north west);
                    \path (A.center) \pgfextra{\pgftext{the man's grandmother said the fire was lit after the suspects attempted to carjack her grandson, \emph{entity33} reported}};}
                  \\
                  she said:' he said he was pumping gas and some guys came up and asked for the car\\
                  ' they pulled out a gun and he took off running\\
                  ' they took the gas tank and started spraying\\
                  ' no one was injured during the fire , but the car 's entire front end was torched , according to \emph{entity52}
                  \\
                  \tikz[baseline=(A.base)]{\node[opacity=0](A) {the \emph{entity53} is investigating the incident as an arson and the suspects remain at large};
                    \shade[inner color=red!20] (A.south east) rectangle (A.north west);
                    \path (A.center) \pgfextra{\pgftext{the \emph{entity53} is investigating the incident as an arson and the suspects remain at large}};}
                  \\
                  surveillance video of the incident is being used in the investigation\\
                  before the fire , which occurred at 12:15am on Saturday , the suspects tried to carjack the man hot case\\
                  the \emph{entity53} is investigating the incident at the \emph{entity67} station as an arson\\
                  \hline
                  \textbf{word extraction}: \\
                  gang poured gasoline in the car, \textit{entity5} Saturday morning.
                  the driver argued with the suspects.
                  his grandmother said the fire was lit by the suspects attempted to carjack her grandson. \\ \hline
                  \textbf{entities}:\\
                  \emph{entity5}:California  \, \emph{entity13}:76-Station \, \emph{entity14}: South LA \, \emph{entity16}:Dodge Charger \, \emph{entity33}:ABC \, \emph{entity52}:NBC \, \emph{entity53}:LACFD \, \emph{entity67}:LA76\\
                  \hline
		\end{tabular}
                \caption{Visualization of the summaries for a
                  DailyMail article. The top half shows the relative
                  attention weights given by the sentence extraction
                  model. Darkness indicates sentence importance. The
                  lower half shows the summary generated by the word
                  extraction.}% Post-processing were performed to
%                  recover the stemmed words and remove redundant
%                  punctuations. }
		\label{vis}
	\vspace*{-2.5mm}
	\end{figure*}
	
\section{Results}
\label{sec:results}
	
Table~\ref{duc} (upper half) summarizes our results on the DUC 2002 test
dataset using \textsc{Rouge}. \mbox{\textsc{nn-se}} represents our
neural sentence extraction model, \mbox{\textsc{nn-we}} our word
extraction model, and \textsc{nn-abs} the neural abstractive
baseline. The table also includes results for the \textsc{lead}
baseline, the logistic regression classifier (\textsc{lreg}), and
three previously published systems (\textsc{ilp}, \textsc{tgraph}, and
\textsc{urank}).
	
\begin{table}[t]
\begin{center}
\begin{small}
\begin{tabular}{|@{~}l@{~}|@{\hspace{.8em}}r@{\hspace{.8em}}r@{\hspace{.8em}}r@{\hspace{.8em}}r@{\hspace{.8em}}r@{\hspace{.8em}}r@{~}|@{~}c@{~}|} \hline
Models & 1$^{st}$ &  2$^{nd}$ & 3$^{rd}$ &4$^{th}$ &5$^{th}$ &	6$^{th}$ & MeanR\\ \hline \hline
\textsc{lead} & 0.10 & 0.17 & 0.37 & 0.15 & 0.16 & 0.05 & 3.27\\
\textsc{ilp} & 0.19 & 0.38 & 0.13  &  0.13  & 0.11  & 0.06 & 2.77\\
\textsc{nn-se} & 0.22 & 0.28 &  0.21 & 0.14 & 0.12 &  0.03 & 2.74 \\
\textsc{nn-we} & 0.00 & 0.04 & 0.03 &  0.21 & 0.51 & 0.20 & 4.79 \\
\textsc{nn-abs} & 0.00   & 0.01  & 0.05  & 0.16  &  0.23 & 0.54 & 5.24\\
Human & 0.27 & 0.23 & 0.29  & 0.17 & 0.03  & 0.01  & 2.51 \\ \hline
\end{tabular}
\end{small}
\end{center}
\caption{\label{tab:humans} Rankings (shown as proportions) and mean
  ranks  given to systems by human participants (lower is better). 
}
\vspace*{-2.5mm}
\end{table}
	
The \mbox{\textsc{nn-se}} outperforms the \textsc{lead} and
\textsc{lreg} baselines with a significant margin, while performing
slightly better than the \textsc{ilp} model. This is an encouraging
result since our model has only access to embedding features obtained
from raw text.  In comparison, \textsc{lreg} uses a set of manually
selected features, while the \textsc{ilp} system takes advantage of
syntactic information and extracts summaries subject to
well-engineered linguistic constraints, which are not available to our
models. Overall, our sentence extraction model achieves performance
comparable to the state of the art without sophisticated constraint
optimization (\textsc{ilp}, \textsc{tgraph}) or sentence ranking
mechanisms (\textsc{urank}). We visualize the sentence weights of the
\mbox{\textsc{nn-se}} model in the top half of Figure~\ref{vis}. As
can be seen, the model is able to locate text portions which
contribute most to the overall meaning of the document.

\textsc{Rouge} scores for the word extraction model are less
promising. This is somewhat expected given that \textsc{Rouge} is
$n$-gram based and not very well suited to measuring summaries which
contain a significant amount of paraphrasing and may deviate from the
reference even though they express similar meaning. However, a
meaningful comparison can be carried out between \textsc{nn-we} and
\textsc{nn-abs} which are similar in spirit.  We
observe that \textsc{nn-we} consistently outperforms the purely
abstractive model. As \mbox{\textsc{nn-we}} generates summaries by
picking words from the original document, decoding is easier for this
model compared to \mbox{\textsc{nn-abs}} which deals with an open
vocabulary.  The extraction-based generation approach is more robust
for proper nouns and rare words, which pose a serious problem to open
vocabulary models.  An example of the generated summaries for
\textsc{nn-we} is shown at the lower half of Figure~\ref{vis}.

Table~\ref{duc} (lower half) shows system results on the 500
DailyMail news articles (test set). 
In general, we observe similar
trends to DUC~2002, with \mbox{\textsc{nn-se}} performing the best in
terms of all \textsc{rouge} metrics. Note that scores here are
generally lower compared to DUC~2002. This is due to the fact that the
gold standard summaries (aka highlights) tend to be more laconic and
as a result involve a substantial amount of paraphrasing. More experimental results on this dataset are provided in the appendix.

The results of our human evaluation study are shown in
Table~\ref{tab:humans}. Specifically, we show, proportionally, how
often our participants ranked each system 1st, 2nd, and so on.  Perhaps
unsurprisingly, the human-written descriptions were considered best
and ranked 1st 27\% of the time, however closely followed by our
\textsc{nn-se} model which was ranked 1st 22\% of the time. The
\textsc{ilp} system was mostly ranked in 2nd place (38\% of the
time). The rest of the systems occupied lower ranks. We further
converted the ranks to ratings on a scale of 1 to 6 (assigning ratings
6$\dots$1 to rank placements 1$\dots$6).  This allowed us to perform
Analysis of Variance (ANOVA) which revealed a reliable effect of
system type. Specifically, post-hoc Tukey tests showed that
\textsc{nn-se} and \textsc{ilp} are significantly (\mbox{$p < 0.01$})
better than \textsc{lead}, \textsc{nn-we}, and \textsc{nn-abs} but do
not differ significantly from each other or the human goldstandard.

\section{Conclusions}
\label{sec:conclusions}
	
In this work we presented a data-driven summarization framework based
on an encoder-extractor architecture. We developed two classes of
models based on sentence and word extraction. Our models can be
trained on large scale datasets and learn informativeness features
based on continuous representations without recourse to linguistic
annotations. Two important ideas behind our work are the creation of
hierarchical neural structures that reflect the nature of the
summarization task and generation by extraction. The later effectively
enables us to sidestep the difficulties of generating under a large
vocabulary, essentially covering the entire dataset, with many
low-frequency words and named entities.

Directions for future work are many and varied. One way to improve the
word-based model would be to take structural information into account
during generation, e.g., by combining it with a tree-based algorithm
\cite{cohn2009sentence}.  It would also be interesting to apply the
neural models presented here in a phrase-based setting similar to
\newcite{lebret2015phrase}. A third direction would be to adopt an
information theoretic perspective and devise a purely unsupervised
approach that selects summary sentences and words so as to minimize
information loss, a task possibly achievable with the dataset created
in this work.

\section*{Acknowledgments}We would like to thank three anonymous reviewers and 
members of the ILCC
at the School of Informatics for their valuable feedback. The support
of the European Research Council under award number 681760
``Translating Multiple Modalities into Text'' is gratefully
acknowledged.

	%\section*{Acknowledgments}
\section{Appendix}
In addition to the DUC 2002 and 500 DailyMail samples, we additionally report results on the entire DailyMail test set (Table \ref{ducap}). Since there is no established evaluation standard for this task, we experimented with three different ROUGE limits: 75 bytes, 275 bytes and full length. 

\begin{table}[t]
	%		\small
	\centering
	\begin{tabular}{|@{~}l@{\hspace{2ex}}|c@{\hspace{2ex}}c@{\hspace{2ex}}c@{~}|} \hline
		\multicolumn{1}{|@{~}l|}{DM 75b} &
		\textsc{Rouge-1}  & \textsc{Rouge-2}  & \textsc{Rouge-l} \\ \hline\hline
		\textsc{Lead} & 21.9 &  7.2 & 11.6 \\
		\textsc{nn-se} & 22.7 & 8.5 & 12.5\\
		\textsc{nn-we} &  16.0 & 6.4 & 10.2 \\
		%			\raisebox{.1cm}[0pt]{\begin{sideways}DailyMail\end{sideways}}
		\hline 
		\multicolumn{4}{c}{} \\\hline
		\multicolumn{1}{|@{~}l|}{DM 275b} &
		\textsc{Rouge-1}  & \textsc{Rouge-2}  & \textsc{Rouge-l} \\ \hline\hline
		\textsc{Lead} & 40.5 &  14.9 & 32.6 \\
		\textsc{nn-se} & 42.2 & 17.3 & 34.8\\
		\textsc{nn-we} & 33.9 & 10.2 & 23.5 \\
		%			\raisebox{.1cm}[0pt]{\begin{sideways}DailyMail\end{sideways}}
		\hline 
		\multicolumn{4}{c}{} \\\hline
		\multicolumn{1}{|@{~}l|}{DM full} &
		\textsc{Rouge-1}  & \textsc{Rouge-2}  & \textsc{Rouge-l} \\ \hline\hline
		\textsc{Lead} & 53.5 &  21.7 & 48.5 \\
		\textsc{nn-se} & 56.0 & 24.9 & 50.2\\
		\textsc{nn-we} & - & - & -\\
		\hline
		%			\raisebox{.1cm}[0pt]{\begin{sideways}DailyMail\end{sideways}}
	\end{tabular}
	\caption{\label{ducap} \textsc{Rouge} evaluation (\%) on the entire 500 DailyMail samples, with different length limits.}
\end{table}
	% include your own bib file like this:
	\bibliographystyle{acl2016}
	\bibliography{acl2016}

\end{document}